\documentclass[preprint,12pt]{elsarticle}

\usepackage{amssymb}
\usepackage{amsthm}
\usepackage{url}

\usepackage{lineno}

\usepackage{graphics} 
\usepackage{epsfig} 
\usepackage{mathptmx} 
\usepackage{times} 
\usepackage{amsmath} 
\usepackage{amssymb} 
\usepackage{hyperref}
\usepackage{multirow}
\usepackage{diagbox}
\usepackage{gensymb}

\usepackage{algorithm}
\usepackage[noend]{algpseudocode}
\usepackage{pifont}
\usepackage{tikz}
\usepackage{mathtools}

 \newcommand{\qlanda}[0]{{$Q(\lambda)$}}

\journal{Computer Vision and Image Understanding}

\begin{document}

\begin{frontmatter}


\title{Deep Active Object Recognition by Joint Label and Action Prediction}

\author[mm]{Mohsen Malmir\corref{cor1}}
\ead{mmalmir@eng.ucsd.edu}

\author[ks]{Karan Sikka}
\ead{ksikka@eng.ucsd.edu}

\author[df]{Deborah Forster}
\ead{forster@ucsd.edu}

\author[if]{Ian Fasel}
\ead{ian@emotient.com}

\author[if]{Javier R. Movellan}
\ead{javier@emotient.com}

\author[mm]{Garrison W. Cottrell}
\ead{gary@eng.ucsd.edu}

\cortext[cor1]{Corresponding Author}

\address[mm]{Computer Science and Engineering Department, University of California San Diego, 9500 Gilman dr., San Diego CA 92093, USA}
\address[ks]{Electrical and Computer Engineering Department, University of California San Diego, 9500 Gilman dr., San Diego CA 92093, USA}

\address[df]{Qualcomm Inst., University of California San Diego, 9500 Gilman dr., San Diego CA 92093, USA}

\address[if]{Emotient.com, 4435 Eastgate Mall, Suite 320, San Diego, CA, USA, San Diego CA 92121, USA}

\begin{abstract}
An active object recognition system has the advantage of being able to act in the environment to capture images that are more suited for training and that lead to better performance at test time. In this paper, we propose a deep convolutional neural network for active object recognition that simultaneously predicts the object label, and selects the next action to perform on the object with the aim of improving recognition performance. We treat active object recognition as a reinforcement learning problem and derive the cost function to train the network for joint prediction of the object label and the action. A generative model of object similarities based on the Dirichlet distribution is proposed and embedded in the network for encoding the state of the system. The training is carried out by simultaneously minimizing the label and action prediction errors using gradient descent. We empirically show that the proposed network is able to predict both the object label and the actions on GERMS, a dataset for active object recognition. We compare the test label prediction accuracy of the proposed model with Dirichlet and Naive Bayes state encoding. The results of experiments suggest that the proposed model equipped with Dirichlet state encoding is superior in performance, and selects images that lead to better training and higher accuracy of label prediction at test time.  
 \end{abstract}


\end{frontmatter}


\section{Introduction}
\label{sec:intro}

\par{A robot interacting with its environment can collect large volumes of dynamic sensory input to overcome many challenges presented by static data. A robot  manipulating an object with the capability to control its camera orientation, for example, is an example of an active object recognition system. In such dynamic interactions, the robot can select the training data for its models of the environment, with the goal of maximizing the accuracy with which it perceives its surroundings. In this paper, we focus on active object recognition (AOR) with the goal of developing a model that can be used by a robot to recognize an object held in its hand.}

\par{There are a variety of approaches to active object recognition, the goal of which is to  re-position sensors or change the environment so that the new inputs to the system become less ambiguous for label prediction \cite{Aloimonos1988,Bajcsy1988,Denzler2001}. 
An issue with previous approaches to active object recognition is that they mostly used small simplistic datasets, which were not reflective of challenges in real-world applications \cite{mmalmir2015}. To avoid this problem, we have collected a large dataset for active object recognition, called GERMS\footnote{Available at \url{http://rubi.ucsd.edu/GERMS/}}, which contains more than 120K high resolution (1920x1080) RGB images  of 136 different plush toys. This paper extends our previous work, Deep Q-learning \cite{mmalmir2015}, where an action selection network was trained on top of a pre-trained convolutional neural network. In this paper we extend the model to train the network end-to-end using GERMS images to jointly predict object labels and action values.}

\par{This paper makes two primary contributions: First, we develop a deep active object recognition (DAOR) model to jointly predict the label and the best next action on an input image. We propose a deep convolutional neural network that outputs the object labels and action-values in different layers of the network. We use reinforcement learning to teach the network to predict the action values, and minimize the action value prediction error along with the label prediction cross entropy. The visual features in early stages of this network are learned to minimize both errors. The second contribution of this work is to embed a generative Dirichlet model of objects similarities for encoding the state of the system. This model integrates information from different images into a  vector, based on which actions are calculated to optimize performance. We embed this model as a layer in the network and derive the learning rule for updating the Dirichlet parameters using gradient descent. We conduct a series of experiments on the GERMS dataset to test (1) if the model can be trained jointly for label and action prediction, and (2) how effective is the proposed Dirichlet state encoding compared to more traditional Naive Bayes approach, and (3) discuss some of the properties of the learned policies.}
%
%
%

\par{
In the next section, we review some of the previous approaches to active object recognition and examine the datasets they used. Next we introduce the GERMS dataset and describe the training and testing data used for the experiments in this paper. After that, we describe the details of the proposed network and Dirichlet state encoding, going into the details of cost functions and update rules for different layers of the network. In the results section, we report the properties of the proposed network and compare its performance in different scenarios. The following section is the concluding remarks.  
}

%
%
%
%
%

\section{Literature Review}

\par{Active object recognition systems include two modules: A recognition module and a control module. Given a sequence of images, the recognition module produces a belief state about the objects that generated those images. Given this belief state, the control module produces actions that will affect the images observed in the future \cite{Denzler2001}. The controller is typically designed to improve the speed and accuracy of the recognition module.}
\par{One of the earliest active systems for object recognition was developed by Wilkes and Tsotsos \cite{Wilkes1992}. They used a heuristic procedure to bring the object into a \lq standard\rq \, view by a robotic-arm-mounted camera. In a series of experiments on 8 Origami objects, they qualitatively report promising results for achieving the standard view and retrieving the correct object labels. Seibert and Waxman explicitly model the views of an object by clustering the images acquired from the view-sphere of the object into aspects \cite{Seibert1992}. The correlation matrices between these aspects are then used in an aspect network to predict the correct object label. Using three model aircraft objects, they show that the belief over the correct object improves with the number of observed transitions compared to randomly generated paths on the view sphere of these objects.}
%
%
%

\par{Schiele and Crowley developed a framework for active object recognition by making an analogy between object recognition and information transmission \cite{Schiele1998}. They try to minimize the conditional entropy $H(O|M)$ between the original object $O$ and the observed signal $M$. They used the COIL-100 dataset for their experiments, which consists of 7200 images of 100 toy objects rotated in depth \cite{Nene1996}. This dataset has been appealing for active object recognition because it provides systematically defined views of objects. At test time, by sequentially moving to the most and second most discriminative views of each object, Schiele and Crowley achieved almost perfect recognition accuracy on this dataset.}

\par{Borotschnig et al. formulate the observation planning in terms of maximization of the expected entropy loss over actions \cite{Borotschnig2000}. Larger entropy loss is equivalent to less ambiguity in interpreting the image. With an active vision system consisting of a turntable and a moving camera, they report improvements in object recognition over random selection of the next viewing pose on a small set of objects. 
Callari and Ferrie take into account the object modeling error and search for actions that simultaneously minimize both modeling variance and uncertainty of belief over objects \cite{Callari2001}. Using a set of 10 custom clay objects, they report decrease in the entropy of the classifier output and Kullback-Leibler divergence between the posterior distribution of each object and the corresponding true distribution.}

\par{Browatzki et al. use a particle filter approach to determine the viewing pose of an object held in-hand by an iCub humanoid robot \cite{Browatzki2012,Browatzki2014}. For selecting the next best action, instead of maximizing the expected information gain, which is computationally expensive, they maximize a measure of variance of observations across different objects.
They show that their method is superior to random action selection on small sets of custom objects. Atanasov et al. focus on the comparison of myopic greedy action selection that looks ahead only one step and non-myopic action selection which considers several time steps into the future \cite{Atanasov2014}. They formulate the problem as a Partially Observable Markov Decision Process, showing their method is superior to random and greedy selection of actions on a small set of household objects.}

\par{Rebguns et al. used acoustic properties of objects to learn an infomax controller to recognize a set of 10 objects \cite{Rebguns2011}. In this work, they proposed a Dirichlet based model to fuse information from different observations into a single belief vector. Using this latent variable mixture model for acoustic similarities, the robot learned to rapidly reduce uncertainty about the categories of the objects in a room. The state encoding of our system is similar to the mixture model of this work, however we embed this model into the network and train its parameters using gradient descent which is more suited for neural networks.}
%
%
%

\par{Paletta \& Pinz \cite{Paletta2000} treat active object recognition as an instance of a reinforcement learning problem, using Q-learning to find the optimal policy. They used an RBF neural network with the reward function  depending on the amount of entropy loss between the current and the next state.}

\par{
A common trend in many of these approaches is the use of small, sometimes custom- designed sets of objects. There are medium sized datasets such as COIL-100, which consists of 7200 images of 100 toy objects rotated in depth \cite{Nene1996}. This dataset is not an adequately challenging dataset for several reasons, including the simplicity of the image background, and the high similarity of different views of the objects due to single-track recording sessions. What is missing is a challenging dataset for active object recognition with inherent similarities among different object categories. The dataset should be large enough to train models with large number of parameters, such as deep convolutional neural networks. In the next section, we describe GERMS, a large and challenging dataset for active object recognition that we used for experiments in this paper.
%
%
%
}

\section{The GERMS Dataset}
\par
\begin{figure*}[ht!]
   \centering
   \includegraphics[width=.95\textwidth]{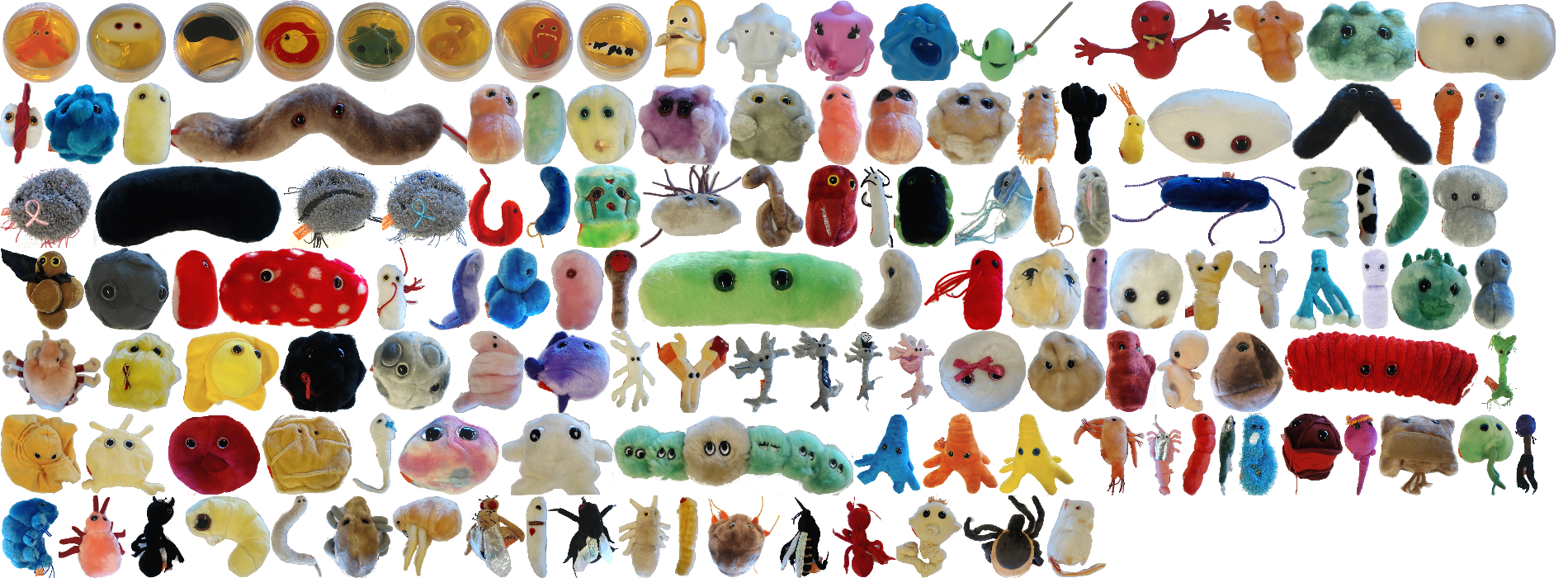}

   \caption{The GERMS dataset. The objects represent human cell types, microbes and disease-related organisms.}
   \label{figure:collage}
  \end{figure*}

{The GERMS dataset was collected in the context of the RUBI project, whose goal is to develop robots that interact with toddlers in early childhood education environments \cite{Malmir2012,Movellan2013, mmalmir2015}. This dataset consists of 1365 video recordings of give-and-take trials using 136 different objects. The objects are soft toys depicting various human cell types, microbes and disease-related organisms. Figure \ref{figure:collage} shows the entire set of these toys. Each video consists of the robot (RUBI) bringing the grasped object to its center of view, rotating it by $180\degree$ and then returning it. The dataset was recorded from RUBI's head-mounted camera at 30 frames per second.}

\par{
The data for GERMS were collected in two days. On the first day, each object was handed to RUBI in one of 6 pre-determined poses, 3 to each arm, after which RUBI grabbed the object and captured images while rotating it. The robot also captured the positions of its joints for every capture image. On the second day, we asked a set of human subjects to hand the GERM objects to RUBI in poses they considered natural. A total of 12 subjects participated in test data collection, each subject handing between 10 and 17 objects to RUBI. For each object, at least 4 different test poses were captured. The background of the GERMS dataset was provided by a large screen TV displaying video scenes from the classroom in which RUBI operates, including toddlers and adults moving around.
}

\par{We use half of the data collected in day 1 and 2 for training and the other half of each day for testing. More specifically, three random tracks out of six tracks for each object in Day 1 and two randomly selected tracks for each object from Day 2 were used for training the network and the rest was used for testing. Table \ref{table:datasetstats} shows the statistics of training and testing data for the experiments in this paper.
}


\begin{table}
\small
\caption{GERMS dataset statistics (mean$\pm$std)}
\label{table:datasetstats}
\begin{center}
\begin{tabular}{|p{1.cm}|p{2cm}|p{2cm}|p{2.25cm}|}
\hline
\textbf{} & \textbf{Number of tracks} & \textbf{Images per Track}& \textbf{Total Number of Images} \\
\hline
Day 1 & $816$  & $157\pm12$ & 76,722 \\
\hline
Day 2 & $549$  & $145\pm19$ & 51,561\\
\hline
\end{tabular}
\end{center}
\end{table}

\section{Proposed Network}
\par{The traditional view of an active object recognition pipeline usually treats the visual recognition and action learning problems separately, with visual features being fixed when learning actions. In this work, we try to solve both problems simultaneously to reduce the training time of an AOR model. By incorporating the errors from action prediction into visual feature extraction, we hope to acquire features that are suited for both label and action prediction.}

\par
\begin{figure*}[ht!]
  \centering
       \includegraphics[width=1.\textwidth]{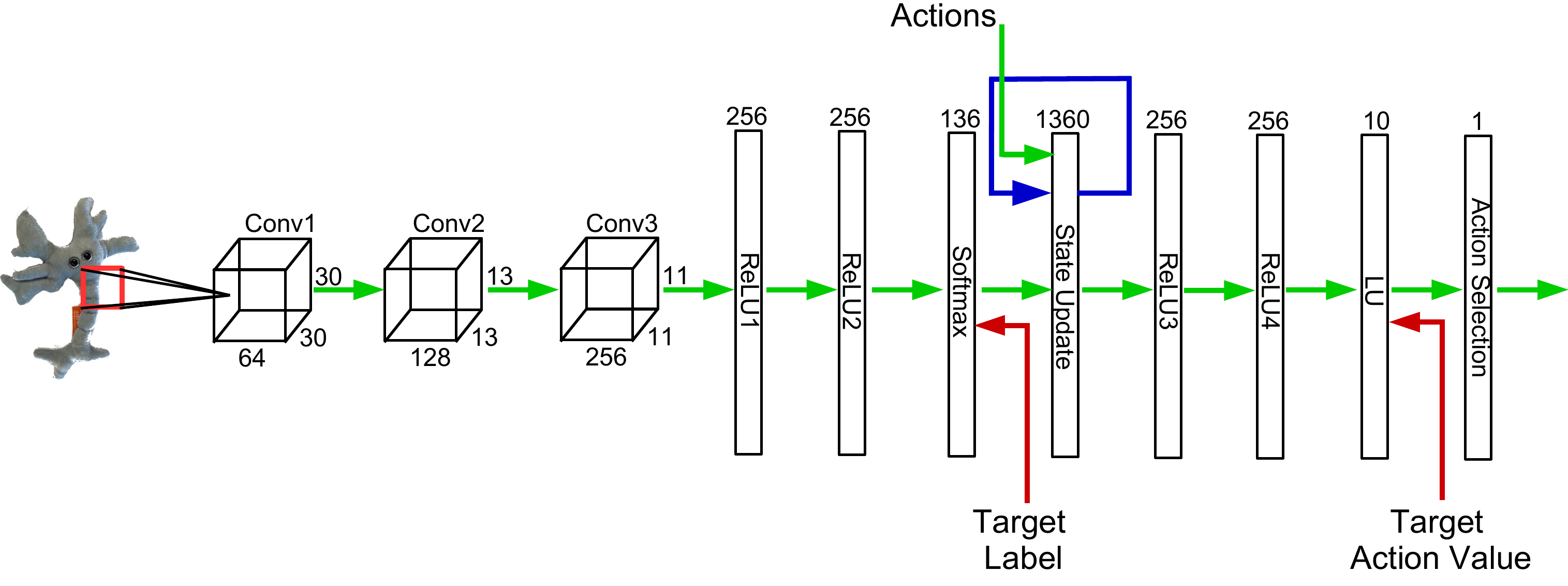}
    \caption{The proposed network for active object recognition. The red arrows representing the target values indicate the layer at which the target values are used to train the network. The numbers represent the number of units in each layer of the network. See table \ref{table:numparams} for more details.}
 \label{figure:architecture}
\end{figure*}

{The proposed network is shown in figure \ref{figure:architecture}. The input image is first transformed to a set of beliefs over different object labels by a classification network. The belief is then combined with the the previously belief vectors to produce an encoding of the \emph{state} of the system. This is done by the \emph{Mixture belief update} layer in the network. The accumulated belief is then transformed into action-values, that are used to select the next input image.}

\par{We next detail each part of the network, describing the challenges and their corresponding solution. We first address the transformation of images into beliefs over object classes. Then we tackle belief accumulation over observed images,followed by the action learning and,finally, present the full description of the algorithm to train this model.}

\subsection{Single Image Classification}
\par{The goal of this part of the network is to transform a single image into beliefs over different object labels. The feature extraction stage is comprised of 3 convolution layers followed by 3 fully connected layers. The dimensions of each layer are shown in figure \ref{figure:architecture}. The convolution layers use filters of $3\times 7\times7$, $64\times 5\times5$ and $128\times 3\times3$ respectively for layers 1,2 and 3. The number of parameters in each layer of the network is shown in table \ref{table:numparams}. The operations of each layer are inspired by the model proposed in \cite{alexnet}. Each convolution layer is followed by rectification, normalization across channels and max pooling over a neighborhood of $2\times 2$ with stride of 1. The dropout for ReLU1 and ReLU2 uses $P=0.5$.} 
\par{
We shall denote the GERMS dataset by $D=\{I_i,y_i,P_i\}_{i=1}^{N}$, where $I_i \in \mathbb{R}^{64 \times 64 \times 3} $ is the image captured by the robot camera, $y_i \in \{o_1,o_2,...,o_c\}$ is the object label and $P_i$ is an integer number denoting the pose of the robot's gripper as positive integers \cite{mmalmir2015}. In order to learn the weights of the single image classification part, we perform gradient decent on action prediction and cross-entropy costs, denoted by $\mathbb{C}_{RL}$ and $\mathbb{C}_{CL}$ respectively. The cross-entropy classification cost $\mathbb{C}_{CL}$ is:
\begin{align}
\label{eq:costCL}
\mathbb{C}_{CL} = - \sum_{i} \sum_{j=1}^C \mathbb{I}(y_i=c) \log B_{ij}.
\end{align}
Here $\mathbb{I}$ is the indicator function for the class of the object and $B_{ij} = P(o_j|I_i)$ is the predicted label belief for the $i^{th}$ image corresponding to the $j^{th}$ class. The next subsection describes the action prediction cost $\mathbb{C}_{RL}$. 
}

\begin{table}[ht!]
\caption{Number of units and parameters for the proposed network.}
\label{table:numparams}
\begin{center}
\begin{tabular}{|c|c|c|c|}

\hline
\textbf{Layer} & \textbf{Number of Units} & \textbf{Input to Unit} & \textbf{Num. Parameters}  \\
\hline
Conv1 & 64x30x30 & $3\times7\times7$ & 9K \\
\hline
Conv2  & 128x13x13 & $64\times5\times5$ & 204K\\
\hline
Conv3  & 256x11x11 & $128\times3\times3$ & 294K\\
\hline
ReLU1  & 256 & 30976 & 7M \\
\hline
ReLU2  & 256 & 256 & 65K \\
\hline
Softmax  & 136 & 256 &  34K \\
\hline
State Update.  & 1360 & 136 & 184K \\
\hline
ReLU3  & 256 & 1360+256 &  413K \\
\hline
ReLU4  & 256 & 256 & 65K \\
\hline
LU  & 10 & 256 & 2K \\
\hline

\end{tabular}
\end{center}
\vspace{-0.2in}
\end{table}

\subsection{Action Value Prediction}

\par{Active object recognition can be treated as a reinforcement learning problem, whose goal is to learn an optimal policy $\pi^*: S \rightarrow A$ from states $S$ to actions $A$. The optimal policy is expected to maximize the total reward for every \emph{interaction sequence} $s^\pi_{0:T}$ with the environment,
\[
s_0 \xrightarrow[]{\pi(s_0)} s_1 \xrightarrow[]{\pi(s_1)} s_2 \xrightarrow[]{\pi(s_2)} \ldots \xrightarrow[]{\pi(s_{T--1})} s_T
\] 
where $s_i \xrightarrow[]{\pi(s_i)} s_{i+1}$ is the transition from $s_i$ to $s_{i+1}$ by performing the action $a_i = \pi(s_i)$. The \emph{total reward} for an interaction sequence $s^\pi_{0:T}$ is $TR(s^\pi_{0:T}) = \sum_{t=0}^T \gamma^t R(s_t)$ where $R:S \rightarrow \mathbb{R}$ is a reward function and $\gamma,\;\; 0 < \gamma < 1$ is a discount factor used to emphasize rewards closer in time. For an AOR system, an interaction sequence starts by observing image of the object with the initial orientation in the robot's gripper. The state of the system is then updated by the observed image, and an action is selected to perform on the object to maximize the total reward. The reward in each step is determined by the accuracy of predicted label for the observed images up to that step.}

\par{  
In order to learn the optimal policy, we use the \qlanda algorithm to train the network to predict actions for improved classification \cite{watkins1989}. This is a model-free method that learns to predict the expected reward of actions in each state. More specifically, let $Q^\pi(s,a)$ be the \emph{action value} for state $s$ and action $a$,
\[
Q^\pi(s,a) = E_{\pi} \left \{ TR(s^\pi_{0:T})  | s_0=s, a_0=a \right  \},
\]
which is the expected reward for doing action $a$ in state $s$. Let the agent interact with the environment to produce a set of interaction sequences $\{ s^\pi_{0:T} \}$. Then \qlanda learns a policy by applying the following update rule to every observed transition $TR^{\pi}(s_t,s_{t+1}) = s_t \xrightarrow[]{\pi(s_t)} s_{t+1}$,
\begin{align}
Q^\pi(s_t,a_t) \leftarrow (1-\alpha) Q^\pi(s_t,a_t) + \alpha \left [ R(s_{t+1}) + \gamma \max_{a} Q^\pi(s_{t+1},a) \right ]
\end{align}
where $0 < \alpha < 1$ is the learning rate, and action $a_t$ is selected using an epsilon-greedy version of the learned policy. We interpret this iterative update in the following way to be useful for training a neural network. Let the output layer of the network predict $Q(s,a)$ for the learned policy $\pi$ for every possible action $a$ in $s$. Then a practical approximation of the optimal policy is obtained by minimizing the reinforcement learning cost,
\begin{align}
\label{COSTRL} \mathbb{C}_{RL} = \sum_{TR^{\pi}(s_t,s_{t+1}) \in \{ s^\pi_{0:T}\} } \left [  R(s_{t+1}) + \gamma \max_{a} Q^\pi(s_{t+1},a) - Q^\pi(s_t,a_t)\right  ]^2
\end{align}
In the proposed network, action value prediction is done by transforming the state of the system $S_t$ at $t^{th}$ through layers ReLU3,ReLU4 and LU2. We train the weights of the network in these layers by minimize $\mathbb{C}_{RL}$. In the next subsection, we go into the details of state encoding, and after that we describe the set of actions. 
}

\subsection{State Encoding}
\par{
State encoding has a prominent effect on the performance of an AOR system. Based on the current state of the system, an action is selected that is expected to decrease the ambiguity about the object label. An appealing choice is to transform images into beliefs over different target classes and use them as the state of the system. Based on the target label beliefs, the system decides to perform an action to improve its target label prediction. What we expect from the AOR system is to guide the robot to pick object views that are more discriminative among target classes. 
}
\par{
We first transform the input image $I_i$ into a belief vector $B_i = [B_{ij}]_{j=1}^C$ using the the first 7 layers of the network, where 
\[
B_{ij} \geq 0, \sum_{j=1}^C B_{ij}=1,
\]
The produced label belief vector is then combined with the previously observed belief vectors from this interaction sequence to form the state of the system. The motivation for this encoding is that the combined belief encodes the ambiguity of the system about target classes and thus can be used to navigate to more discriminative views of objects. Active object recognition methods usually adapt a Naive Bayes approach to combining beliefs from different observations. Assume that in an interaction sequence, a sequence of images $I_{0:t} = \{ I_0, I_1,\ldots,I_t\}$ have been observed and their corresponding beliefs $B_{0:t} = \{ B_0, B_1,\ldots,B_t\}$ have been calculated. The state of the system at time $t$ is calculated using Naive Bayes belief combination, which is to take the product of the individual belief vectors and then normalize,
\begin{align}
\label{ST} s_t = P(O| I_{0:t}) &= \frac{P(O,I_{0:t})} {P(I_{0:t})} \nonumber\\
&\propto P(O) \prod_{i=0}^t P(I_i | O) \nonumber\\
&\propto \prod_{i=0}^t P(O|I_i)
\end{align}
where $O$ is the target label, and $P(O|I_i)$ is the vector of beliefs produced using single image classification. Here we assumed a uniform prior over images and target labels. The problem with Naive Bayes is that if an image is observed repeatedly in $I_{1:t}$, the result will change based on the number of repetitions. This is undesirable since the state of the system changes with repeated observations of an image where no new information is added to the system. If a specific image is good for classification, the system can visit that image more often to artificially increase the performance of the system. To avoid this problem, we adapt a generative model based on Dirichlet distribution to combine different belief vectors.
}
 
\par{
We use a generative model similar to  \cite{Rebguns2011} to calculate the state of the system given a set of images. The intuition behind this model is that performing an action on an object will produce a distribution of belief vectors. We model the observed belief vectors given the object and action as a Dirichlet distribution, parameters of which are learned from the data. The model is shown in figure \ref{fig:Dirichlet}. Here $A$ is a discrete variable representing the action from the repertoire of actions $\{ a^1,a^2,\ldots,a^H\}$, $O \in \{o^1,o^2,\ldots,o^C\}$ represents the object label and $\alpha \in \mathbb{R}^C$ is the vector of parameters of the Dirichlet distribution from which the belief vector $B\in \mathbb{R}^C$ over target labels is drawn,
\begin{align}
P(B | \alpha) &= \text{Dir}(B; \alpha) \nonumber \\
&= \frac{\Gamma(\sum_{j=1}^C [\alpha]_j)}{\prod_{j=1}^C \Gamma([\alpha]_j)} \prod_{j=1}^C [B]_j^{[\alpha]_j-1}
\end{align}
}

\par{
The state of the system is calculated by computing the posterior probability of object-action beliefs using the model in figure \ref{fig:Dirichlet}. Let $P^O_a(a_i,B_i) = P(O,a | a_i,B_i)$ denote the posterior probability of an object-action pair given the performed action and the observed belief vector. Assuming uniform prior over object and $\alpha$ and a deterministic policy for choosing actions,
\begin{align}
P(o,a| B) &= \nonumber\\
&= \frac{\int_\alpha P(O,a,B,\alpha) d\alpha}{P(B)} \nonumber\\
&\propto \int_\alpha P(O) P(a) P(\alpha | O,a)P(B|\alpha) d\alpha \nonumber \\
\label{POab} &\propto \int_{\alpha^O_a} Dir(B ; \alpha^O_a)) d\alpha^O_a
\end{align}
The notation $\alpha^O_a$ is to make clear that there is an $\alpha$ for each pair of object-action.
Instead of full posterior probability, we use $\hat{\alpha}^O_a$, maximum likelihood estimate of $\alpha$, and replace the integral above by ,
\begin{align}
P(o,a|B) \approx Dir(B | \hat{\alpha}^O_a)
\end{align}
For an interaction sequence $B_{0:t}$ and $A_{0:t} = \{ a_0,a_1,\ldots,a_t\}$, the posterior probability of object-action pair is,
\begin{align}
P(O ,a | A_{0:t},B_{0:t}) = \prod_{i=0}^t P(O,a|B_i)^{\mathbb{I}(a,a_i)}
\end{align}
The state of the system is comprised of the vector of object posterior beliefs for every object and action, plus the features and belief extracted from the latest image $I_t$,
\begin{align}
\label{DirichletState}
s_t = \{ [P(O,a | A_{0:t},B_{0:t})] , B_t \}, \\ 
O \in \{o^1,o^2,\ldots,o^C\} \nonumber\\ 
a\in \{a^1,a^2,\ldots,a^H\} \nonumber
\end{align}
Note that $s_t \in \mathbb{R}^{CH}$ is a vector of length $C\times H$.
}

\subsection{Training Network for Joint Label and Action Prediction}
\par{
Our goal is to train the network jointly for action and label prediction. We achieved this by minimizing the total cost which is sum of the costs for both label (\ref{eq:costCL}) and action prediction (\ref{COSTRL}). Note that the errors for action value prediction are backpropagated through the entire network, reaching visual feature extraction units. The total cost function for action value and label prediction is,
\begin{align}
\label{cost} Cost = \mathbb{C}_{RL} + \mathbb{C}_{CL} 
\end{align}

The weights of the network in the visual feature extraction layers (Conv1, Conv2, Conv3, ReLU1, ReLU2, LU1) are trained using backpropagation on (\ref{cost}), while the action prediction layers (ReLU3, ReLU4 and LU2) are trained by gradient descent on the action prediction error (\ref{COSTRL}).
}

\par
\begin{figure*}[t]
 \centering
 \includegraphics[width=0.9\textwidth]{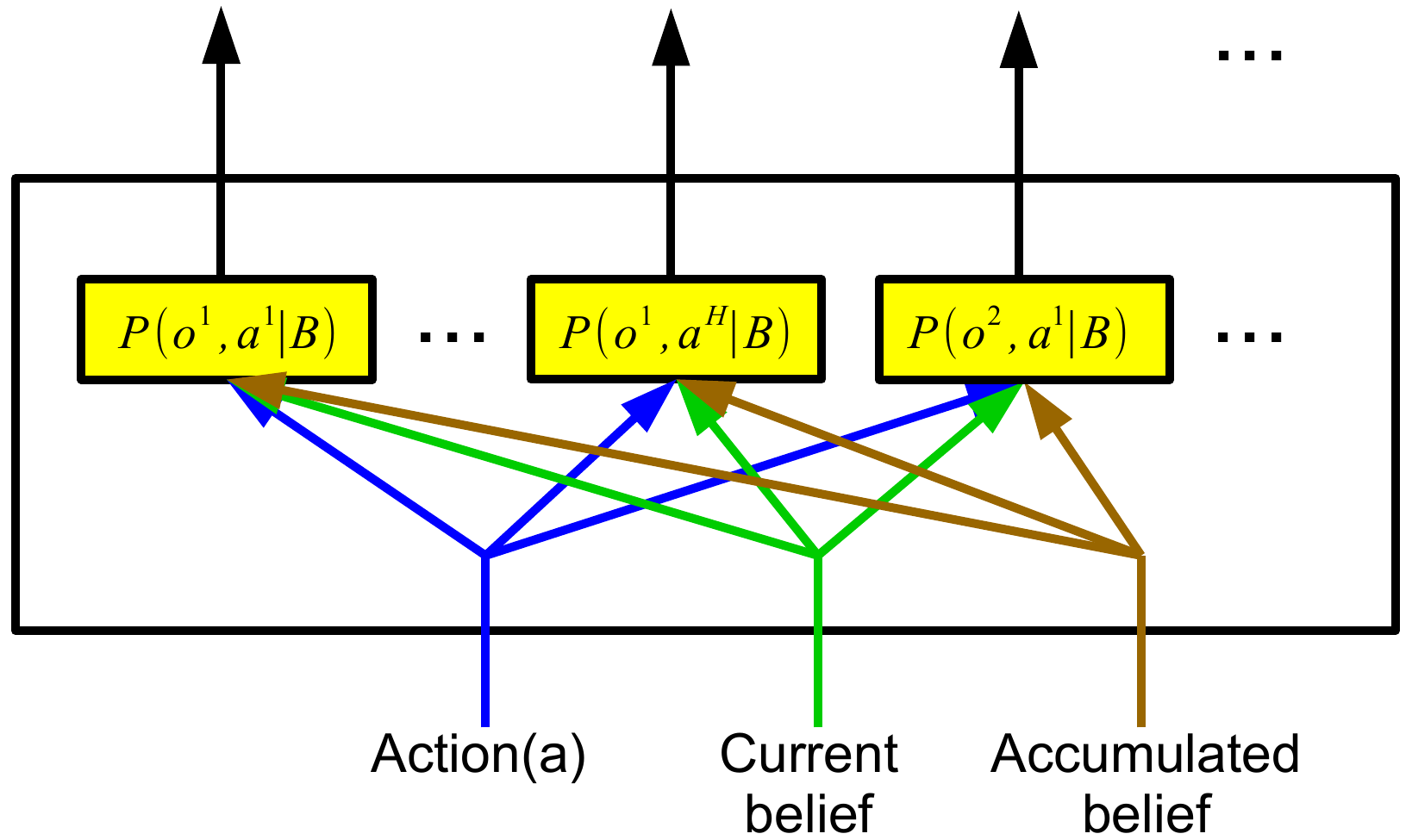}
 \caption{Dirichlet belief update layer. Each unit in this layer represents a Dirichlet distribution for a pair of object-action. The parameters of this layer are the vectors of Dirichlet parameters $\alpha^O_k$ for each unit.}
 \label{fig:Dirichlet}
\end{figure*}

{
To learn the parameters of the belief update, that is $\alpha^O_a$, we use gradient descent on maximum likelihood of the data. The maximum likelihood of Dirichlet distribution is a convex function of its parameters and can be minimized using gradient descent. For a set of beliefs $B_{1:N}$ observed by performing action $a$ on the object $O$, the gradient of the log-likelihood with respect to the parameters are,
\begin{align}
\frac{\partial \log P(B_{1:N} | \alpha^O_a) }{\partial [\alpha^O_a]_k} &= N \frac{d}{d[\alpha^O_a]_k} \log\Gamma(\sum_{j=1}^C [\alpha^O_a]_j) -  \frac{d}{d[\alpha^O_a]_k} \log\Gamma([\alpha^O_a]_k) + \log B_k \nonumber \\ 
\label{costDirichlet} &= N \Psi(\sum_{j=1}^C [\alpha^O_a]_j) -  N \Psi([\alpha^O_a]_k) + \log B_k
\end{align}
where $\Psi(x) = d/d(x) \log \Gamma(x)$ is the digamma function. We use one unit per Dirichlet distribution $Dir(|\alpha^O_a)$ in the belief update. These units receive the current belief and their output for the previously observed belief, and produce an updated belief. An schematic of the belief update layer of the network is shown in figure \ref{fig:Dirichlet}. Learning $\alpha^O_k$ is carried out simultaneously with the rest of the network weights in one training procedure.
}

\subsection{Reward Function}
\par{
Another component that has an important effect on the performance of our AOR system is the reward function which maps state of the system (\ref{ST}) into rewards. A simple choice for reward function is
\begin{align}
\label{RSTClf}
R(s_t) = \left \{ \begin{array}{ll}
+1 & \text{if } \arg\max_i [B_t]_i =  \text{Target-Label}(I_t)) \\
-1 & \text{otherwise}
\end{array} \right .
\end{align}
We call this the \emph{correct-label reward} function. A reward of $+1(-1)$ is given to the system if at time step $t$ the action $a_t$ brings the object to a pose for which the predicted label is correct (wrong). The intention behind this reward function is to drive the AOR system to pick actions that lead to best next view of the object in terms of label prediction. 
}

\subsection{Action Coding}
\par{
In order to be able to reach every position in the robot's joint gripper range, we use a set of relative rotations as the actions of the system. More specifically, we use 10 actions to rotate the gripper from its current position by any of the following offset values: $\{ \pm \frac{\pi}{4},\pm \frac{\pi}{8},\pm \frac{\pi}{16},\pm \frac{\pi}{32},\pm \frac{\pi}{64} \}$. The total range of rotation for each of the robot's grippers is $\pm \pi$. The actions are selected to be fine grained enough so that the robot can reach any position with minimum number of movements possible. This encoding is simple and flexible in the range of positions that the robot can reach, however we found that the policies can become stuck with a few actions without trying the rest. Encoding the states with the Dirichlet belief update helps alleviate this issue to some degree, however, it doesn't completely remove the problem. We deal with this problem by forcing the algorithm to pick the next best action whenever the best action leads to an image which has already been seen.
}

\section{Experimental Results}
\subsection{Training Details}
\par
{
We trained the network by minimizing the costs of classification, action value prediction (\ref{COSTRL}) and negative of log-likelihood of Dirichlet distributions (\ref{costDirichlet}). We used backpropagation with minibatches of size 128 to train the network. For \qlanda, we used initial learning rate of $0.1$ which was multiplied by $0.5$ after iterations $400,800,1200,1500$ and then remained constant. The total number of training iterations is $4000$. For each iteration, an interaction sequence of length 5 is followed. The full training algorithm is shown in algorithm \ref{alg:traindqn}. For \qlanda, we used $\epsilon$-greedy policy in the training stage, with $\epsilon$ decreasing step-wise from 0.9 to 0.1. We found that using an $\epsilon > 0$ at the test stage hurts the performance, therefore we used $\epsilon = 0$ during testing. The number of actions is 10 as described above, and there are a total of 136 object classes, resulting in a total of 1360 Dirichlet distributions for state encoding \ref{DirichletState}.
%
%
%
}

\begin{algorithm}[t!]
\caption{Training the network for joint label and action prediction.}
\label{alg:traindqn}
\begin{algorithmic}[1]
\Procedure{Train}{}
\State $R \leftarrow 1$
\For {iteration=1 To N}
  \State $I_1,y \leftarrow$ \text{NextImage}(\text{iteration})
  \State $s_0 \leftarrow [0]$
  \State $\text{Actions} \leftarrow \text{RandomActions(NumActions)}$
  \For {t=1 To \text{NumMoves}}
      \State $s_t,\text{predictedActions} \leftarrow  \text{FeedForward}(I_t,s_{t-1},\text{Actions}) $
      \State $I_{t+1},\_ \leftarrow \text{NextImage}(I_t,\text{predictedActions})$ 
      \State $\text{targetActionVals},\hat{y} \leftarrow \text{LookAhead}(I_{t+1},s_t,\text{Actions}) $
      \If {$t = \text{NumMoves }$}
    		  	\State $\text{targetActionVals} \leftarrow \text{targetActionVals} +  R(s_t)$
      \EndIf
      \For{ $W \in \{ReLU3,ReLU4,LU\}$}
      		\State $W \leftarrow W - \lambda_W \frac{\partial}{\partial W} \{\mathbb{C}_{RL}\}$
      \EndFor
      \For{ $W \in \{Conv1,Conv2,Conv3,ReLU1,ReLU2,Softmax\}$}
      		\State $W \leftarrow W - \lambda_W \frac{\partial}{\partial W} \{\mathbb{C}_{RL}+\mathbb{C}_{CL}\}$
      \EndFor
      \For{ $O \in \{o^1,o^2,\ldots,o^C\} , a \in \{a^1,a^2,\ldots,a^H\}$}
      		\State $\alpha^O_a \leftarrow \alpha^O_a + \lambda \frac{\partial}{\partial \alpha^O_a} \log P(B_t | \alpha^O_a)$
      \EndFor
  \EndFor

\EndFor
\EndProcedure
\end{algorithmic}
\end{algorithm}

\subsection{Learning the Parameters of Dirichlet Distributions}
\begin{figure}[t]
 \centering
 \includegraphics[width=0.7\textwidth]{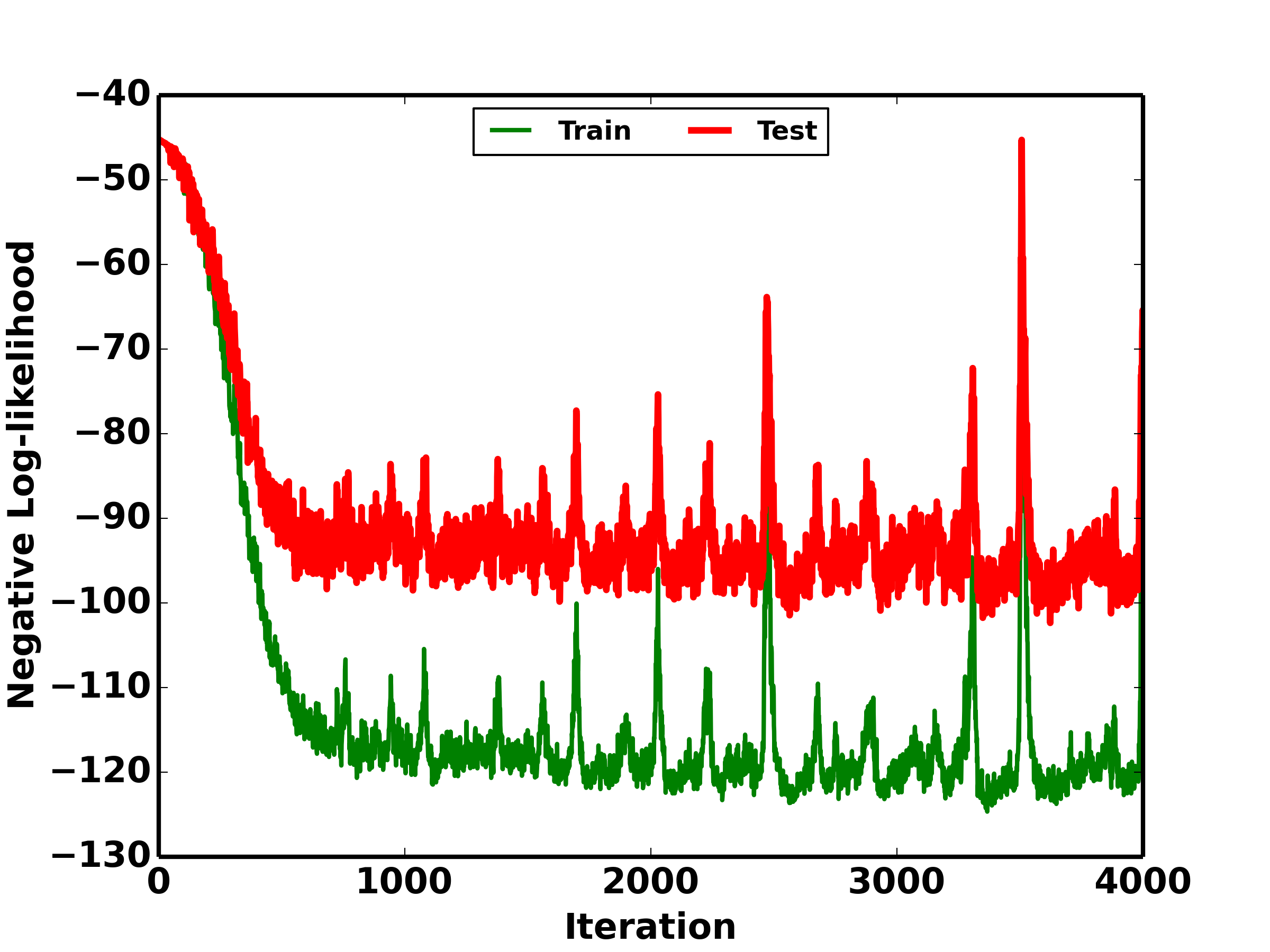}
 \caption{Average Negative log-likelihood of data under Dirichlet distributions. The decrease in negative log-likelihood indicates learning in the belief update layer.}
 \label{fig:lpao}
\end{figure}

Figure \ref{fig:lpao} shows the average negative log-likelihood of the data under Dirichlet distributions for training of a DN model. This figure shows that the neg-log-likelihood of data decreases after the first 1000 iterations, after which is the rate of change is decreased but not stopped. 
%
%
%

\subsection{Label Prediction Accuracy}
\subsubsection{Comparing Naive Bayes and Dirichlet State Encoding}
\par
\begin{figure}[ht!]
  \centering
  \setlength{\unitlength}{\textwidth} 
    \begin{picture}(1,0.5)
       \put(-0.1,0){\includegraphics[width=1.2\unitlength]{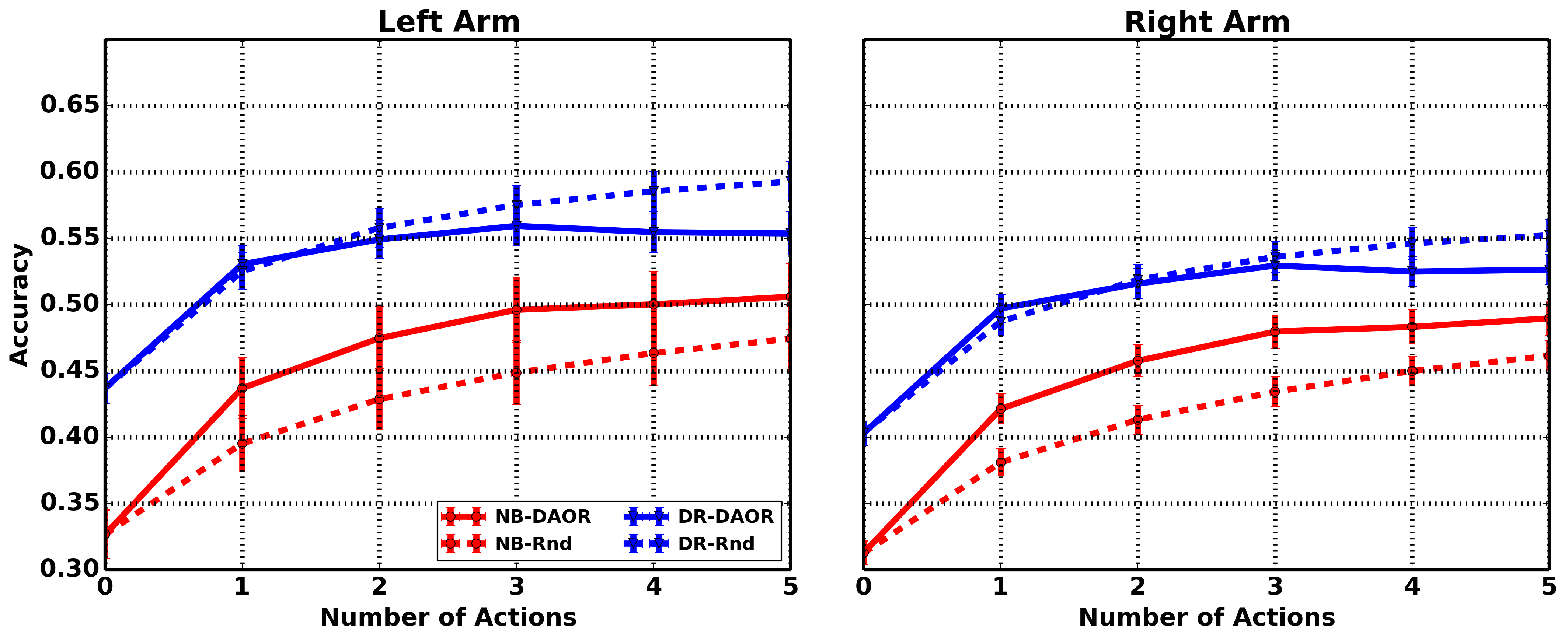}}
    \end{picture}
    \caption{Test label prediction accuracy as a function of number of observed images for left and right arms for Dirichlet state encoding with repeated visits (DR) and non-repeated visits (DN).}
 \label{figure:DRNB}
\end{figure}

{
In the first experiment, we compare the effectiveness of the Dirichlet and Naive Bayes state encodings in terms of label prediction accuracy. For Naive Bayes models (NB), the state of the system is updated using (\ref{ST}), while the size and configuration of the rest of the network remains the same. Dirichlet state encoding is implemented using (\ref{DirichletState}). We refer to Dirichlet models as (DR). For each encoding and for each arm, we train 10 different models and report the average test label prediction accuracy as a function of number of observed images, comparing the Deep Active Object Recognition (DAOR) and Random (Rnd) action selection policies. Figure \ref{figure:DRNB} plots the performance for these models. It is obvious that the Dirichlet model is superior to Naive Bayes in label prediction accuracy. 
%
%
%
}

\par{
The first point to notice in figure \ref{figure:DRNB} is the performance difference between Naive Bayes and Dirichlet belief updates on single images (action 0). NB models achieve a performance less than $35\%$, while Dirichlet achieves higher than $40\%$. One interpretation of this result is that the Naive Bayes models pick actions that bounce between a subset of train images, leading to underfitting of the model. In the next subsection, we provide some evidence for this justification. On the other hand, the performance of DR-DAOR model tends to saturate after 3 actions, while DR-Rnd keeps improving for subsequent actions. This might be due to the fact that DR-DAOR also bounces between subsets of images at the test time. We can avoid such behavior by forcing the policies to pick actions that lead to joint poses that haven't already been visited in the same interaction sequence.}
%
%
%
%
%

\subsubsection{Removing Duplicate Visits }
\par
\begin{figure}[ht!]
  \centering
  \setlength{\unitlength}{\textwidth} 
    \begin{picture}(1,0.5)
       \put(-0.1,0){\includegraphics[width=1.2\unitlength]{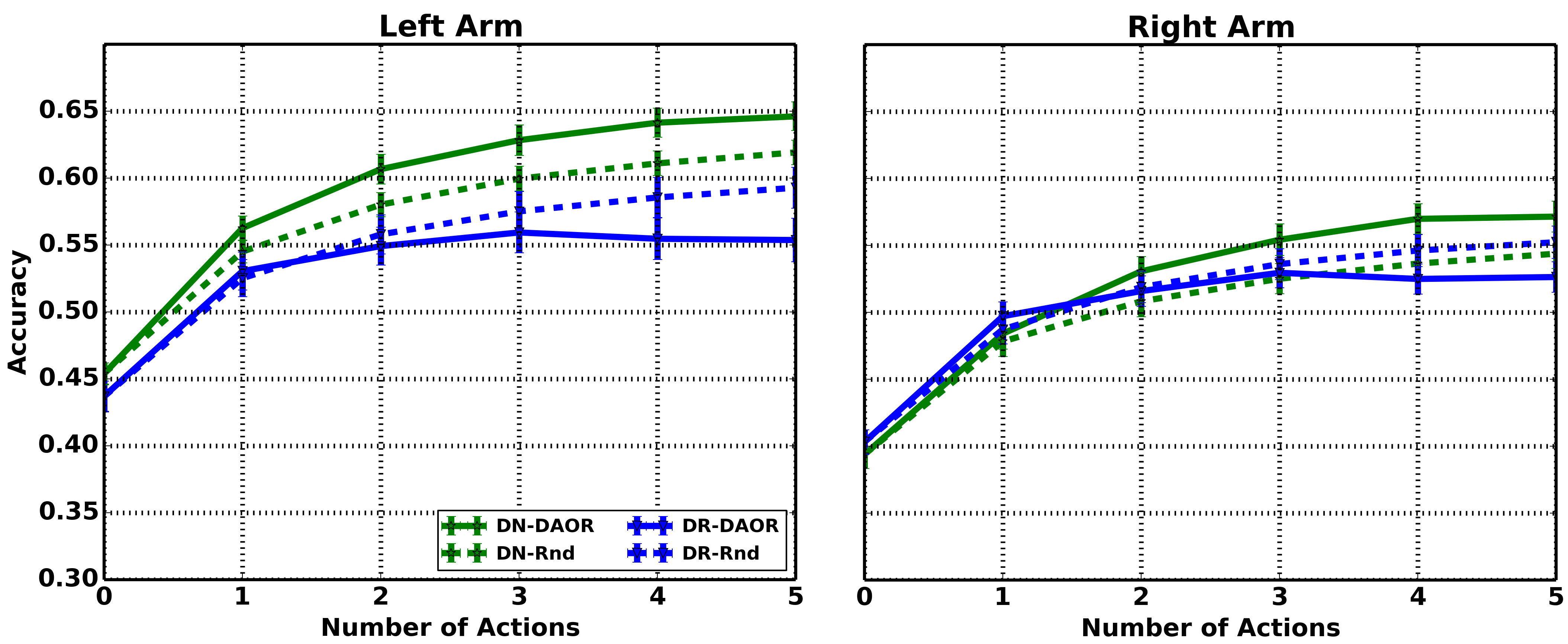}}
    \end{picture}
    \caption{Test label prediction accuracy as a function of number of observed images for left and right arms for Naive Bayes (NB) and Dirichlet (DR) state encoding.}
 \label{figure:DRDN}
\end{figure}

{
We train a set of models using Dirichlet state encoding, while forcing the policy to pick non-duplicate joint poses in every action of an interaction sequence. This approach is easy to implement by keeping a history of visited joint poses during an interaction sequence and picking actions with highest action value that don't lead to the visited joint positions. We refer to this model as Dirichlet with non-repeated visits (DN). Comparison between DN and DR for Rnd and DAOR policies (both forced to visit novel poses) is shown in figure \ref{figure:DRDN}.
}

\par{
Comparison between the models mentioned above is shown in table \ref{table:baseline}. We see that the best performing model is DN-DAOR with the exception of action 1 for the right arm, which DR-DAOR achieves the best performance. For both arms, Dirichlet models perform significantly better than Naive Bayes, improving the model's performance on average by $10\%$ for the right arm and $14\%$ for the left arm.
}

\begin{table}[h!]
\caption{Comparison of DQN, random and sequential.}
\label{table:baseline}
\begin{center}
\begin{tabular}{|l|c|c|c|c|c|c|c|}
\cline{1-7}
\diagbox{Policy}{Observed\\Frames} & 0&1&2&3&4&5 \\
\hline
NB-Rnd&  31.3 &  38.1 &  41.3 &  43.4 & 45.0 & 46.1& \multicolumn{1}{ |c|  }{\multirow{5}{*}{\textbf{Right Arm} }}\\
\cline{1-7}
NB-DAOR&31.3&42.1&45.8&48.0&48.3&49.0&\multicolumn{1}{ |c|  }{}\\
\cline{1-7}
DR-RND&40.3 & 48.7 &  51.9 & 53.6 & 54.6 & 55.2 &\multicolumn{1}{ |c|  }{}\\
\cline{1-7}
DR-DAOR& 40.3 &  \textbf{49.7} & 51.6 & 53.0 & 52.5 & 52.6&\multicolumn{1}{ |c|  }{}\\
\cline{1-7}
DN-RND& 39.4 & 47.8 &  50.8 &  52.5 &  53.6 &  54.3&\multicolumn{1}{ |c|  }{}\\
\cline{1-7}
DN-DAOR& 39.3 &  48.4 &  \textbf{53.1} & \textbf{55.4} & \textbf{57.0} & \textbf{57.1}&\multicolumn{1}{ |c|  }{}\\
\hline
\hline
NB-Rnd& 32.7 & 39.5 & 42.9 & 44.9 & 46.3 & 47.4& \multicolumn{1}{ |c|  }{\multirow{5}{*}{\textbf{Left Arm} }}\\
\cline{1-7}
NB-DAOR& 32.7 & 43.7 & 47.5 & 49.6 & 50.0 & 50.6 &\multicolumn{1}{ |c|  }{}\\
\cline{1-7}
DR-RND& 43.7 &  52.5 & 55.8 & 57.5 & 58.6 & 59.3 &\multicolumn{1}{ |c|  }{}\\
\cline{1-7}
DR-DAOR& 43.7 &  53.0 & 54.9 & 55.9 & 55.5 & 55.4&\multicolumn{1}{ |c|  }{}\\
\cline{1-7}
DN-RND& 45.4 &  54.5 & 58.0 & 60.0 & 61.1 & 61.9&\multicolumn{1}{ |c|  }{}\\
\cline{1-7}
DN-DAOR&45.4 &  \textbf{56.3} & \textbf{60.7} & \textbf{62.8} & \textbf{64.1} & \textbf{64.6} &\multicolumn{1}{ |c|  }{}\\
\hline
\end{tabular}
\end{center}
\vspace{-0.2in}
\end{table}

\subsubsection{Visualizing Policies }
\par
\begin{figure*}[h!]
 \centering
 \includegraphics[width=1.\textwidth]{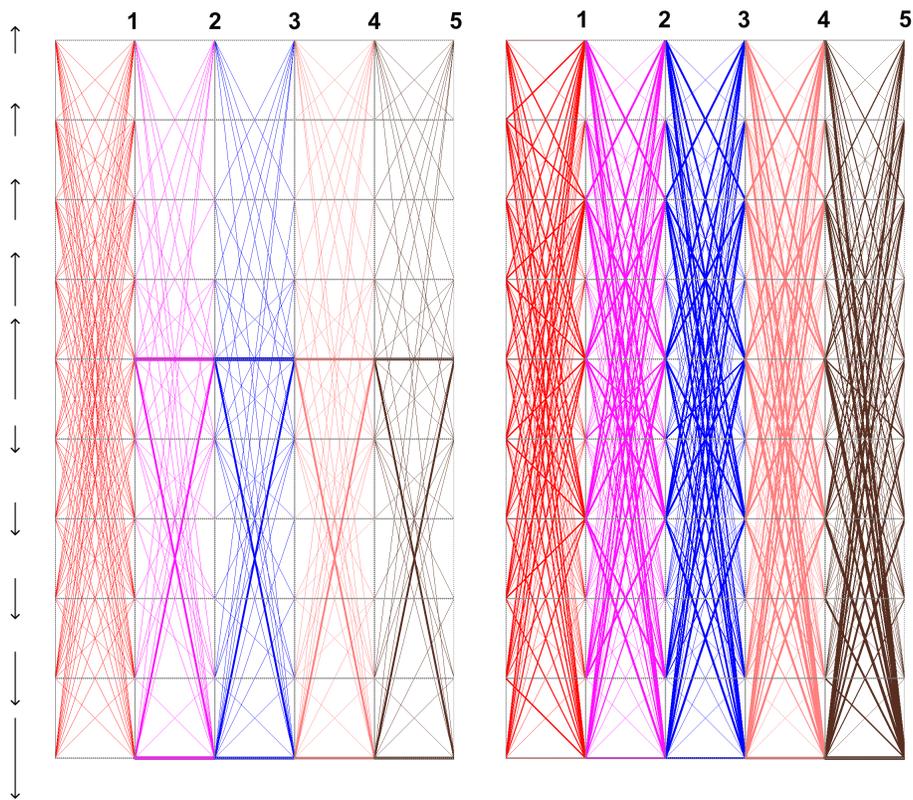}
 \caption{Visualization of (left) NB and (right) DN models for train data. Each row represents an action and each column represents a move performed by the policy in an interaction sequence. The color of lines connecting two columns are different for clarity for every consecutive time steps, while the thickness of these line indicate the frequency of that transition in interaction sequences. }
 \label{figure:policytrain}
\end{figure*}

\begin{figure*}[h!]
 \centering
 \includegraphics[width=1.\textwidth]{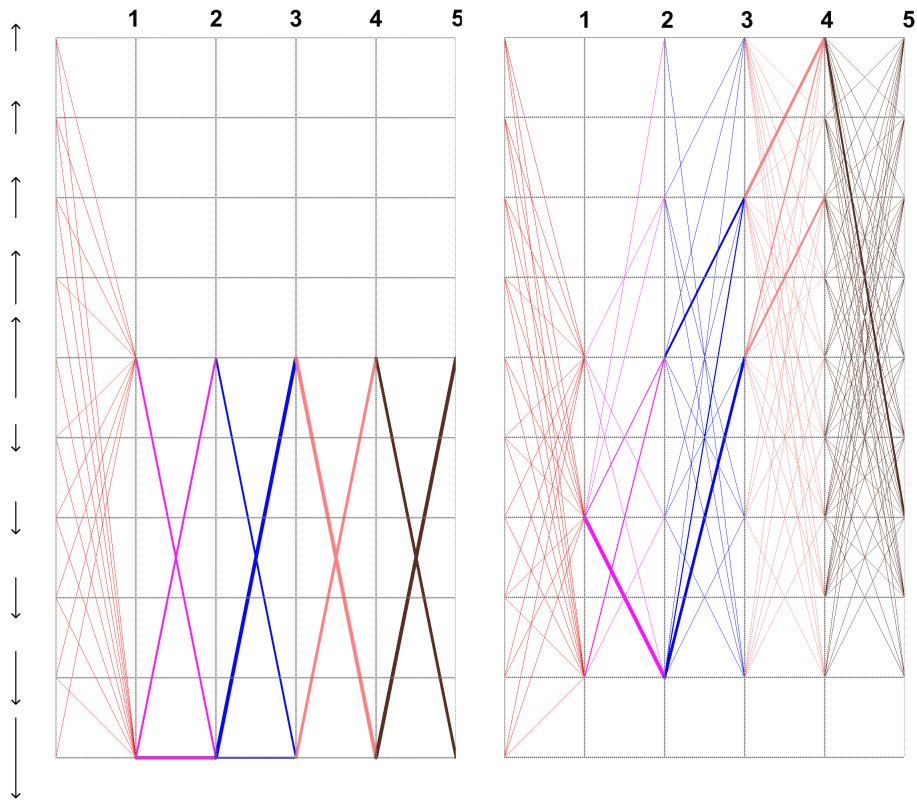}
 \caption{Visualization of (left) NB and (right) DN models for test data. NB model prefers to repeats the same two actions, swinging between two joint poses at one end of the joint range. The DN model usually performs a few larger rotations on the object, followed by a few smaller rotations in different directions.}
 \label{figure:policytest}
\end{figure*}

{
It may help us understand the weakness and strength of different models if we take a closer look into the learned policies. For this purpose, we visualize the consecutive actions in the interactions sequences of length 5, as shown for training data in figures \ref{figure:policytrain} and for test data in figure \ref{figure:policytest}. Each plot represents actions  in different rows, with the magnitude and orientation of the action begin depicted by the length and direction of the corresponding arrow on the left side. Each time step of the interaction sequence is shown as a numbered column. The colored lines in each plot connect one action in column $i$ to another action in column $i+1$ only if those actions appeared consecutively in interaction sequences at these time steps. The thickness of lines depicts the relative frequency by which two actions were observed on the data.
}
\par{
Figure \ref{figure:policytrain} visualizes the policies DN-DAOR and NB-DAOR on the training data. This figure helps clarify the lower performance of NB models as described before. For NB-DAOR shown on the left side of figure \ref{figure:policytrain}, we see thick lines connecting actions that rotate the object with the largest magnitude in opposite directions. The relative thickness of these lines indicates that the model tends to go to one end of joint's rotation range, go back with one large rotation and then repeat. Despite presence of other actions, this back and forth action dominates the training process, leading to lower accuracy on test label prediction for single images. On the right side of figure \ref{figure:policytrain} we see that DN-DAOR picks a wide range of actions, which leads to better examination of training images and thus higher performance on single images.
}

\par{
Figure \ref{figure:policytest} visualizes the learned policies at test time for NB-DAOR and DN-DAOR. We see on the left side that NB-DAOR only swings between the two large rotations in the opposite direction, while DN-DAOR prefers to do a few larger actions (thick purple and blues lines connecting columns 2, 3 and 4) followed by few smaller actions in different directions. There is no back and forth for DN-DAOR between visited joint positions, which leads to better performance on the test set.
}

\section{CONCLUSIONS}

\par{
In this paper, we proposed a model for deep active object recognition based convolutional neural networks. The model is trained by jointly minimizing the action and label prediction simultaneously. The visual features in early stages of this network were trained by minimizing action and label prediction costs. The difference between the work presented here and deeply supervised networks \cite{DeeplySupervised} is that in the latter, the training is carried out by minimizing the classification error in different layers, while in our approach we minimized the action learning costs along with classification error. 
}
\par{
We also adapted an alternative to the common Naive Bayes belief update rule for state encoding of the system. Naive Bayes has the potential to overfit to subsets of training images, which could lead to lower accuracy at the test time. We used a generative model based on Dirichlet distribution to model the belief over target classes and actions performed on them. This model was embedded into the network, which allowed training the network in one pass jointly for label and action prediction. The results of experiments confirmed that the proposed Dirichlet model is superior in test label prediction to the Naive Bayes approach for system's state encoding.
}
\par{
A common trend we observed in the models trained in this paper was the strong preference for a few actions, which led to limited examination of the objects, and thus lower performance on label prediction. This preference was the strongest in the Naive Bayes state encoding models. Employing Dirichlet for state encoding helped alleviate this problem, mainly for the training data and less for test data. We observed that the strong preference for a limited set of actions weakens in the training stage for the DR-DAOR model, and as a result of this the test label prediction accuracy was improved. We hypothesize that in addition to state encoding, learning actions on the training images which have high label prediction accuracy, leads to this strong preference. In training our models, the training accuracy reaches above $90\%$ after 1000 iterations. This may cause the  \qlanda to reward every action, which finally may lead to one action taking over and always producing the highest action value.
}

\section{Acknowledgments}
\par{
The research presented here was funded by NSF IIS 0968573 SoCS, IIS INT2-Large 0808767, and NSF SBE-0542013 and in part by US NSF ACI-1541349 and OCI-1246396, the University of California Office of the President, and the California Institute for Telecommunications and Information Technology (Calit2).
}


\section{References}

\begin{thebibliography}{00}

\bibitem{Aloimonos1988} J. Aloimonos, J. I. Weiss, and A. Bandyopadhyay, Active vision, International J. Computer Vision, vol. 1, no. 4, pp. 333-356, 1988.
\bibitem{Bajcsy1988} R. Bajcsy, Active perception, Proceedings of the IEEE, vol. 76, no. 8, pp. 966-1005, 1988.
\bibitem{Wilkes1992} D. Wilkes and J. K. Tsotsos, Active object recognition, Proceedings CVPR'92., 1992 IEEE Computer Society Conference on, pp. 136-141. IEEE, 1992.


\bibitem{Seibert1992} M. Seibert and A. M. Waxman, Adaptive 3-D object recognition from multiple views, IEEE Transactions on Pattern Analysis and Machine Intelligence, vol. 14, no. 2, pp. 107-124, 1992.
\bibitem{Schiele1998} B. Schiele and J. L. Crowley, Transinformation for active object recognition, In Computer Vision, Sixth International Conference on, pp. 249-254. IEEE, 1998.
\bibitem{Nene1996} S. A. Nene, S. K. Nayar and H. Murase, Columbia object image library (COIL-100), Technical Report CUCS-006-96, Columbia University, 1996.
\bibitem{Borotschnig2000} H. Borotschnig, L. Paletta, M. Prantl and A. Pinz, Appearance-based active object recognition, Image and Vision Computing, vol. 18, no. 9, pp. 715-727, 2000.
\bibitem{Paletta2000} L. Paletta and A. Pinz, Active object recognition by view integration and reinforcement learning, Robotics and Autonomous Systems, vol. 31, no. 1, pp. 71-86, 2000.
\bibitem{Callari2001} F. G. Callari and F. P. Ferrie, Active object recognition: Looking for differences, International J. Computer Vision, vol. 43, no. 3, pp. 189-204, 2001.
\bibitem{Browatzki2012} B. Browatzki, V. Tikhanoff, G. Metta, H. H. Bulthoff and C. Wallraven, Active object recognition on a humanoid robot, In Robotics and Automation (ICRA), 2012 IEEE International Conference on, pp. 2021-2028, IEEE, 2012.
\bibitem{Browatzki2014} B. Browatzki, V. Tikhanoff, G. Metta, H. H. Bulthoff, C. Wallraven, Active In-Hand Object Recognition on a Humanoid Robot, Robotics, IEEE Transactions on , vol. 30, no. 99, pp. 1-9, 2014.
\bibitem{Atanasov2014} N. Atanasov, B. Sankaran, J. L. Ny, G. J. Pappas and K. Daniilidis, Nonmyopic View Planning for Active Object Classification and Pose Estimation, Robotics, IEEE Transactions on , vol. 30, no. 99, pp. 1078-1090, 2014.
\bibitem{Malmir2012} M. Malmir, D. Forster, K. Youngstrom, L. Morrison and J. R. Movellan, Home Alone: Social Robots for Digital Ethnography of Toddler Behavior, Computer Vision Workshops (ICCVW), 2013 IEEE International Conference on, pp. 762-768, 2013.
\bibitem{Movellan2013} J. R. Movellan, M. Malmir and D. Forester, HRI as a tool to monitor socio-emotional development in early childhood education, In proc. HRI 2nd Workshop on Applications for Emotional Robots, Bielefeld, Germany, 2014.


\bibitem{mmalmir2015} M. Malmir, K. Sikka, D. Forster, J. Movellan and G. W. Cottrell. Deep Q-learning for Active Recognition of GERMS: Baseline performance on a standardized dataset for active learning. In Xianghua Xie, Mark W. Jones, and Gary K. L. Tam, editors, Proceedings of the British Machine Vision Conference (BMVC), pages 161.1-161.11. BMVA Press, September 2015.

\bibitem{watkins1989} Watkins, C. J. C. H. (1989). Learning from Delayed Rewards. Ph.D. thesis, Cambridge University.

\bibitem{alexnet} Krizhevsky, Alex, Ilya Sutskever, and Geoffrey E. Hinton. "Imagenet classification with deep convolutional neural networks." In Advances in neural information processing systems, pp. 1097-1105. 2012.

\bibitem{Rebguns2011} Rebguns, Antons, Daniel Ford, and Ian R. Fasel. "Infomax control for acoustic exploration of objects by a mobile robot." In Workshops at the Twenty-Fifth AAAI Conference on Artificial Intelligence. 2011.


\bibitem{Denzler2001}Denzler, Joachim, Christopher M. Brown, and Heinrich Niemann. "Optimal camera parameter selection for state estimation with applications in object recognition." In Pattern Recognition, pp. 305-312. Springer Berlin Heidelberg, 2001.

\bibitem{DeeplySupervised} Chen-Yu Lee, Saining Xie, Patrick Gallagher, Zhengyou Zhang, Zhuowen Tu, Deeply-Supervised Nets, In Proceedings of AISTATS 2015.

\end{thebibliography}
\end{document}